\documentclass[runningheads,a4paper]{llncs}
\usepackage{amsmath}
\usepackage{amssymb}

\usepackage{graphicx}

\usepackage{cite}
\usepackage{nohyperref}  % This makes hyperref commands do nothing without errors
\usepackage{url}  % This makes \url work

\usepackage{color} 
\usepackage{epsfig}
\usepackage{amsfonts}
\usepackage[linesnumbered,vlined,ruled]{algorithm2e}

\def\squareforqed{\hbox{\rlap{$\sqcap$}$\sqcup$}}
\def\qed{\ifmmode\squareforqed\else{\unskip\nobreak\hfil
\penalty50\hskip1em\null\nobreak\hfil\squareforqed
\parfillskip=0pt\finalhyphendemerits=0\endgraf}\fi}
\newcommand{\keywords}[1]{\par\addvspace\baselineskip
\noindent\keywordname\enspace\ignorespaces#1}

\begin{document}

\mainmatter  % start of an individual contribution

% 12 pages maximum for ICONIP 2018!!!

\title{Transductive Learning with String Kernels for Cross-Domain Text Classification\vspace*{-0.2cm}}

\titlerunning{Transductive Learning with String Kernels for Text Classification}

\author{\vspace{-0.2cm}Radu Tudor Ionescu
\and
Andrei M. Butnaru}

\institute{University of Bucharest, 14 Academiei, Bucharest, Romania\\
\email{raducu.ionescu@gmail.com, butnaruandreimadalin@gmail.com}
}

\date{\today}

\toctitle{Transductive Learning with String Kernels for Cross-Domain Text Classification}
\tocauthor{Radu Tudor Ionescu, Andrei M. Butnaru}
\maketitle

%%%%%%%%% ABSTRACT
\begin{abstract}
\vspace{-0,2cm}
For many text classification tasks, there is a major problem posed by the lack of labeled data in a target domain. Although classifiers for a target domain can be trained on labeled text data from a related source domain, the accuracy of such classifiers is usually lower in the cross-domain setting. Recently, string kernels have obtained state-of-the-art results in various text classification tasks such as native language identification or automatic essay scoring. Moreover, classifiers based on string kernels have been found to be robust to the distribution gap between different domains. In this paper, we formally describe an algorithm composed of two simple yet effective transductive learning approaches to further improve the results of string kernels in cross-domain settings. %The first approach is based on interpreting the pairwise string kernel similarities between samples in the training set and samples in the test set as features. Our second approach is a simple self-training method based on two learning iterations. In the first iteration, a classifier is trained on the training set and tested on the test set, as usual. In the second iteration, a number of test samples (to which the classifier associated higher confidence scores) are added to the training set for another round of training. 
By adapting string kernels to the test set without using the ground-truth test labels, we report significantly better accuracy rates in cross-domain English polarity classification.
\vspace*{-0.2cm}
\end{abstract}

\setlength{\abovedisplayskip}{3pt}
\setlength{\belowdisplayskip}{3pt}

%%%%%%%%% BODY TEXT
\section{Introduction}
\vspace*{-0.1cm} 

Domain shift is a fundamental problem in machine learning, that has attracted a lot of attention in the natural language processing and vision communities~\cite{Blitzer-ACL-2007,Pan-WWW-2010,Lui-IJCNLP-2011,Zhuang-IJCAI-2013,Luo-EMNLP-2015,Shu-CIKM-2015,Fernandez-JAIR-2016,Sener-NIPS-2016,Sun-AAAI-2016,Chang-AAAI-2017,franco-EACL-2017}. To understand and address this problem, generated by the lack of labeled data in a target domain, researchers have studied the behavior of machine learning methods in cross-domain settings~\cite{Lui-IJCNLP-2011,Franco-KBS-2015,franco-EACL-2017} and came up with various domain adaptation techniques~\cite{Long-KDE-2014,Shu-CIKM-2015,Fernandez-JAIR-2016,Chang-AAAI-2017}. In cross-domain classification, a classifier is trained on data from a source domain and tested on data from a (different) target domain. The accuracy of machine learning methods is usually lower in the cross-domain setting, due to the distribution gap between different domains. However, researchers proposed several domain adaptation techniques by using the unlabeled test data to obtain better performance~\cite{Joachims-ICML-1999,Ifrim-PKDD-2006,Ceci-ICDMW-2008,Guo-ICDM-2012,Sener-NIPS-2016}. Interestingly, some recent works~\cite{franco-EACL-2017,Radu-Andrei-ADI-2017} indicate that string kernels can yield robust results in the cross-domain setting without any domain adaptation. In fact, methods based on string kernels have demonstrated impressive results in various text classification tasks ranging from native language identification \cite{popescu-ionescu:2013:BEA8,ionescu-popescu-cahill-EMNLP-2014,ionescu-popescu-cahill-COLI-2016,Ionescu-BEA-2017} and authorship identification \cite{PopescuG12} to dialect identification \cite{Ionescu-VarDial-2016,Radu-Andrei-ADI-2017,Ionescu-VarDial-2018}, sentiment analysis \cite{franco-EACL-2017,marius-KES-2017} and automatic essay scoring \cite{Cozma-ACL-2018}. As long as a labeled training set is available, string kernels can reach state-of-the-art results in various languages including English \cite{ionescu-popescu-cahill-EMNLP-2014,franco-EACL-2017,Cozma-ACL-2018}, Arabic \cite{Radu-ICONIP-2015,ionescu-popescu-cahill-COLI-2016,Radu-Andrei-ADI-2017,Ionescu-VarDial-2018}, Chinese \cite{marius-KES-2017} and Norwegian \cite{ionescu-popescu-cahill-COLI-2016}. Different from all these recent approaches, we use unlabeled data from the test set in a transductive setting in order to significantly increase the performance of string kernels. In our recent work~\cite{Ionescu-EMNLP-2018}, we proposed two transductive learning approaches combined into a unified framework that improves the results of string kernels in two different tasks. In this paper, we provide a formal and detailed description of our transductive algorithm and present results in cross-domain English polarity classification.

The paper is organized as follows. Related work on cross-domain text classification and string kernels is presented in Section~\ref{sec_Related_Work}. Section~\ref{sec_String_Kernels} presents our approach to obtain domain adapted string kernels. The transductive transfer learning method is described in Section~\ref{sec_TTL}. The polarity classification experiments are presented in Section~\ref{sec_Polarity_Experiments}. %, and the Arabic dialect identification experiments are presented in Section~\ref{sec_Dialect_Experiments}. 
Finally, we draw conclusions and discuss future work in Section~\ref{sec_Conclusion}.
 
 \vspace*{-0.1cm}
\section{Related Work}
\label{sec_Related_Work}

\vspace*{-0.1cm}
\subsection{Cross-Domain Classification}

Transfer learning (or domain adaptation) aims at building effective classifiers for a target domain when the only available labeled training data belongs to a different (source) domain. Domain adaptation techniques can be roughly divided into graph-based methods~\cite{Pan-WWW-2010,Ponomareva-RANLP-2013,Chang-AAAI-2017,Arun-SDM-2017}, probabilistic models~\cite{Zhuang-IJCAI-2013,Luo-EMNLP-2015}, knowledge-based models~\cite{Ifrim-PKDD-2006,Bollegala-KDE-2013,Franco-KBS-2015} and joint optimization frameworks~\cite{Long-KDE-2014}. The transfer learning methods from the literature show promising results in a variety of real-world applications, such as image classification~\cite{Long-KDE-2014}, text classification~\cite{Joachims-ICML-1999,Guo-ICDM-2012,Zhuang-IJCAI-2013}, polarity classification~\cite{Pan-WWW-2010,Ponomareva-RANLP-2013,Luo-EMNLP-2015,Fernandez-JAIR-2016,Arun-SDM-2017} and others~\cite{Daume-ACL-2007}.

\noindent
{\bf General transfer learning approaches.}
Long et al. \cite{Long-KDE-2014} proposed a novel transfer learning framework to model distribution adaptation and label propagation in a unified way, based on the structural risk minimization principle and the regularization theory.
Shu et al. \cite{Shu-CIKM-2015} proposed a method that bridges the distribution gap between the source domain and the target domain through affinity learning, by exploiting the existence of a subset of data points in the target domain that are distributed similarly to the data points in the source domain.
In \cite{Sener-NIPS-2016}, deep learning is employed to jointly optimize the representation, the cross-domain transformation and the target label inference in an end-to-end fashion. %The authors obtained state-of-the-art results in digit and object recognition from images.
More recently, Sun et al. \cite{Sun-AAAI-2016} proposed an unsupervised domain adaptation method that minimizes the domain shift by aligning the second-order statistics of source and target distributions, without requiring any target labels.
Chang et al. \cite{Chang-AAAI-2017} proposed a framework based on using a parallel corpus to calibrate domain-specific kernels into a unified kernel for leveraging graph-based label propagation between domains. % They presented transfer learning results in cross-language sentiment classification and digit classification.

\noindent
{\bf Cross-domain text classification.}
Joachims~\cite{Joachims-ICML-1999} introduced the Transductive Support Vector Machines (TSVM) framework for text classification, which takes into account a particular test set and tries to minimize the error rate for those particular test samples.
Ifrim et al.~\cite{Ifrim-PKDD-2006} presented a transductive learning approach for text classification based on combining latent variable models for decomposing the topic-word space into topic-concept and concept-word spaces, and explicit knowledge models with named concepts for populating latent variables.
%\newcite{Cohen-ICDMW-2007} presented a maximum entropy-based technique, achieving comparable performance with the TSVM approach~\cite{Joachims-ICML-1999}.
%A hierarchical text classifier that works in the transductive setting is described in \cite{Ceci-ICDMW-2008}.
%\newcite{Lui-IJCNLP-2011} presented a general-purpose language identification system based on a feature selection method that generalizes across domains.
Guo et al.~\cite{Guo-ICDM-2012} proposed a transductive subspace representation learning method to address domain adaptation for cross-lingual text classification.
%\newcite{Li-AAAI-2012} described an approach for cross-domain text classification, that extracts both the shared and the domain-specific latent features to facilitate effective knowledge transfer.
Zhuang et al.~\cite{Zhuang-IJCAI-2013} presented a probabilistic model, by which both the shared and distinct concepts in different domains can be learned by the Expectation-Maximization process which optimizes the data likelihood.
In \cite{Bhatt-CONLL-2015}, an algorithm to adapt a classification model by iteratively learning domain-specific features from the unlabeled test data is described.

\noindent
{\bf Cross-domain polarity classification.}
In recent years, cross-domain sentiment (polarity) classification has gained popularity due to the advances in domain adaptation on one side, and to the abundance of documents from various domains available on the Web, expressing positive or negative opinion, on the other side. 
Some of the general domain adaptation frameworks have been applied to polarity classification~\cite{Zhuang-IJCAI-2013,Bhatt-CONLL-2015,Chang-AAAI-2017}, but there are some approaches that have been specifically designed for the cross-domain sentiment classification task~\cite{Blitzer-ACL-2007,Li-SIGIR-2009,Pan-WWW-2010,Ponomareva-RANLP-2013,Franco-KBS-2015,Luo-EMNLP-2015,Fernandez-JAIR-2016,franco-EACL-2017,Arun-SDM-2017}.
To the best of our knowledge, Blitzer et al.~\cite{Blitzer-ACL-2007} were the first to report results on cross-domain classification proposing the structural correspondence learning (SCL) method, and its variant based on mutual information (SCL-MI).
Pan et al.~\cite{Pan-WWW-2010} proposed a spectral feature alignment (SFA) algorithm to align domain-specific words from different domains into unified clusters, using domain-independent words as a bridge.
% Results on Amazon Reviews, popular method used as baseline
Bollegala et al.~\cite{Bollegala-KDE-2013} used a cross-domain lexicon creation to generate a sentiment-sensitive thesaurus (SST) that groups different words expressing the same sentiment, using unigram and bigram features as~\cite{Blitzer-ACL-2007,Pan-WWW-2010}. % This approach also obtained competitive results in single-domain polarity classification.
%\newcite{Zhou-AAAI-2014} presented a deep learning approach to learn a feature mapping between cross-domain heterogeneous features and a better feature representation for the mapped data in order to reduce the bias issue caused by the cross-domain correspondences.
Luo et al.~\cite{Luo-EMNLP-2015} proposed a cross-domain sentiment classification framework based on a probabilistic model of the author's emotion state when writing. An Expectation-Maximization algorithm is then employed to solve the maximum likelihood problem and to obtain a latent emotion distribution of the author.
% Results on amazon Reviews, only one single domain above 80%.
Franco-Salvador et al.~\cite{Franco-KBS-2015} combined various recent and knowledge-based approaches using a meta-learning scheme (KE-Meta). They performed cross-domain polarity classification without employing any domain adaptation technique.
% Reports results on Amazon reviews, good state-of-the-art
More recently, Fern\'{a}ndez et al.~\cite{Fernandez-JAIR-2016} introduced the Distributional Correspondence Indexing (DCI) method for domain adaptation in sentiment classification. The approach builds term representations in a vector space common to both domains where each dimension reflects its distributional correspondence to a highly predictive term that behaves similarly across domains.
% Results on Amazon reviews
A graph-based approach for sentiment classification that models the relatedness of different domains based on shared users and keywords is proposed in \cite{Arun-SDM-2017}. %The approach models the relatedness of different domains based on shared users and keywords.

\vspace*{-0.1cm}
\subsection{String Kernels}

In recent years, methods based on string kernels have demonstrated remarkable performance in various text classification tasks~\cite{LodhiSSCW02,Escalante-ACL-2011,PopescuG12,ionescu-popescu-cahill-EMNLP-2014,franco-EACL-2017,Radu-Andrei-ADI-2017,Cozma-ACL-2018}. String kernels represent a way of using information at the character level by measuring the similarity of strings through character n-grams. %They are a particular case of the more general convolution kernels~\cite{haussler-1999}. 
Lodhi et al.~\cite{LodhiSSCW02} used string kernels for document categorization, obtaining very good results. String kernels were also successfully used in authorship identification~\cite{PopescuG12}. More recently, various combinations of string kernels reached state-of-the-art accuracy rates in native language identification~\cite{ionescu-popescu-cahill-EMNLP-2014} and Arabic dialect identification~\cite{Radu-Andrei-ADI-2017}. 
Interestingly, string kernels have been used in cross-domain settings without any domain adaptation, obtaining impressive results. For instance, Ionescu et al.~\cite{ionescu-popescu-cahill-EMNLP-2014} have employed string kernels in a cross-corpus (and implicitly cross-domain) native language identification experiment, improving the state-of-the-art accuracy by a remarkable $32.3\%$. 
Gim\'{e}nez-P\'{e}rez et al.~\cite{franco-EACL-2017} have used string kernels for single-source and multi-source polarity classification. Remarkably, they obtain state-of-the-art performance without using knowledge from the target domain, which indicates that string kernels provide robust results in the cross-domain setting without any domain adaptation.
% In~\cite{franco-EACL-2017}, string kernels have successfully been used for sentiment analysis in single-source and multi-source cross-domain settings. The authors report state-of-the-art performance without any domain adaptation. 
Ionescu et al.~\cite{Radu-Andrei-ADI-2017} obtained the best performance in the Arabic Dialect Identification Shared Task of the 2017 VarDial Evaluation Campaign~\cite{dsl2017}, with an improvement of $4.6\%$ over the second-best method. It is important to note that the training and the test speech samples prepared for the shared task were recorded in different setups~\cite{dsl2017}, or in other words, the training and the test sets are drawn from different distributions. Different from all these recent approaches~\cite{ionescu-popescu-cahill-EMNLP-2014,franco-EACL-2017,Radu-Andrei-ADI-2017}, we use unlabeled data from the target domain to significantly increase the performance of string kernels in cross-domain text classification, particularly in English polarity classification. %An attempt to use unlabeled data to improve the string kernel framework for single-domain bio-relation extraction was made by \newcite{Kuksa-ECML-2010}. The authors modified the string kernel framework by implementing an abstraction step, which groups similar words to generate more abstract entities in a semi-supervised setting. Different from their approach, we propose a transductive transfer learning approach suitable for the cross-domain setting. Another difference is that the kernels used in \cite{Kuksa-ECML-2010} are based on word n-grams, while our string kernels are based on character n-grams.
% They employ unsupervised models to capture contextual semantic similarities between words from a large unlabeled corpus. Only work that discusses semi-supervised string kernels.

\vspace*{-0.1cm} 
\section{Transductive String Kernels}
\label{sec_String_Kernels}
\vspace*{-0.1cm} 

\noindent
{\bf String kernels.}
Kernel functions \cite{taylor-Cristianini-cup-2004} capture the intuitive notion of similarity between objects in a specific domain. For example, in text mining, string kernels can be used to measure the pairwise similarity between text samples, simply based on character n-grams. Various string kernel functions have been proposed to date \cite{LodhiSSCW02,taylor-Cristianini-cup-2004,ionescu-popescu-cahill-EMNLP-2014}. Perhaps one of the most recently introduced string kernels is the histogram intersection string kernel \cite{ionescu-popescu-cahill-EMNLP-2014}. For two strings over an alphabet $\Sigma$, $x,y \in \Sigma^*$, the intersection string kernel is formally defined as follows:
\begin{equation}
\begin{split}
k^{\cap}(x,y)=\sum\limits_{v \in \Sigma^p} \min \lbrace \mbox{num}_v(x), \mbox{num}_v(y) \rbrace ,
\end{split}
\end{equation}
where $\mbox{num}_v(x)$ is the number of occurrences of n-gram $v$ as a substring in $x$, and $p$ is the length of $v$. The spectrum string kernel or the presence bits string kernel can be defined in a similar fashion \cite{ionescu-popescu-cahill-EMNLP-2014}.

\noindent
{\bf Transductive string kernels.}
We present a simple and straightforward approach to produce a transductive similarity measure suitable for strings. We take the following steps to derive transductive string kernels. For a given kernel (similarity) function $k$, we first build the full kernel matrix $K$, by including the pairwise similarities of samples from both the train and the test sets. For a training set $X = \{x_1, x_2, ..., x_m\}$ of $m$ samples and a test set $Y = \{y_1, y_2, ..., y_n\}$ of $n$ samples, such that $X \cap Y = \emptyset$, each component in the full kernel matrix is defined as follows:
\begin{equation}
\begin{split}
K_{ij}= k(z_i, z_j),
\end{split}
\end{equation}
where $z_i$ and $z_j$ are samples from the set $Z = X \cup Y = \{x_1, x_2, ..., x_m, y_1, y_2, ..., y_n\}$, for all $1 \leq i,j \leq m + n$. We then normalize the kernel matrix by dividing each component by the square root of the product of the two corresponding diagonal components:
\begin{equation}\label{eq_Kernel_Matrix_Normalization}
\begin{split}
\hat{K}_{ij} = \frac{K_{ij}}{\sqrt{K_{ii} \cdot K_{jj}}}.
\end{split}
\end{equation}
% \hat{K}_{ij} = \frac{K_{ij}}{\sqrt{K_{ii} \cdot K_{jj}}}, \forall 1 \leq i,j \leq m + n.

We transform the normalized kernel matrix into a radial basis function (RBF) kernel matrix as follows:
%\begin{equation}\label{eq_RBF_Kernel}
%\begin{split}
%\tilde{K}_{ij} = exp \left( - \frac{\displaystyle 1 - \hat{K}_{ij}} {\displaystyle 2 \sigma^2} \right).
%\end{split}
%\end{equation}
%% \tilde{K}_{ij} = exp \left( - \frac{\displaystyle 1 - \hat{K}_{ij}} {\displaystyle 2 \sigma^2} \right), \forall 1 \leq i,j \leq m + n.
%As the kernel matrix is already normalized, we can choose $\sigma^2 = 0.5$ for simplicity. Therefore, Equation~\eqref{eq_RBF_Kernel} becomes:
\begin{equation}\label{eq_RBF_Kernel_simple}
\begin{split}
\tilde{K}_{ij} = exp \left(-1 + \hat{K}_{ij}\right).
\end{split}
\end{equation}
% \tilde{K}_{ij} = exp \left(-1 + \hat{K}_{ij}\right), \forall 1 \leq i,j \leq m + n.
Each row in the RBF kernel matrix $\tilde{K}$ is now interpreted as a feature vector. In other words, each sample $z_i$ is represented by a feature vector that contains the similarity between the respective sample $z_i$ and all the samples in $Z$. Since $Z$ includes the test samples as well, the feature vector is inherently adapted to the test set. Indeed, it is easy to see that the features will be different if we choose to apply the string kernel approach on a set of test samples $Y'$, such that $Y' \neq Y$. It is important to note that through the features, the subsequent classifier will have some information about the test samples at training time. More specifically, the feature vector conveys information about how similar is every test sample to every training sample.
We next consider the linear kernel, which is given by the scalar product between the new feature vectors. To obtain the final linear kernel matrix, we simply need to compute the product between the RBF kernel matrix and its transpose:
\begin{equation}
\ddot{K} = \tilde{K} \cdot \tilde{K}'.
\end{equation}
In this way, the samples from the test set, which are included in $Z$, are used to obtain new (transductive) string kernels that are adapted to the test set at hand.

\begin{algorithm}[!tpb]
\small{
\caption{Transductive Kernel Algorithm\label{alg_TTL}}

\textbf{Input}: 

$\mathcal{X} = (X, T) = \lbrace (x_i, t_i) \, \mid \, x_i \in \mathbb{R}^q, t_i \in \{1, 2, ..., c \}, i \in \{1,2, ..., m\} \rbrace$ -- the training set of $m$ training samples and associated class labels;

$Y = \lbrace y_i \, \mid \, y_i \in \mathbb{R}^q, i \in \{1,2, ...,n\} \rbrace$ -- the set of $n$ test samples;

$k$ -- a kernel function;

$r$ -- the number of test samples to be added in the second round of training;

$\mathcal{C}$ -- a binary kernel classifier.

\BlankLine
\textbf{Domain-Adapted Kernel Matrix Computation Steps}:

$Z \leftarrow \{x_1, x_2, ..., x_m, y_1, y_2, ..., y_n\}$\;
$K \leftarrow \boldsymbol{0}_{m+n}$; 
$\hat{K} \leftarrow \boldsymbol{0}_{m+n}$; 
$\tilde{K} \leftarrow \boldsymbol{0}_{m+n}$; 
$\ddot{K} \leftarrow \boldsymbol{0}_{m+n}$\;

\For{$z_i \in Z$}
{
	\For{$z_j \in Z$}
	{
		$K_{ij} \leftarrow k(z_i, z_j)$\;
	}
}

\For{$i \in \{1,2,...,m+n\}$}
{
	\For{$j \in \{1,2,...,m+n\}$}
	{
		$\hat{K}_{ij} \leftarrow \frac{K_{ij}}{\sqrt{K_{ii} \cdot K_{jj}}}$\;
		
		$\tilde{K}_{ij} \leftarrow exp \left(-1 + \hat{K}_{ij}\right)$\;
	}
}

$\ddot{K} = \tilde{K} \cdot \tilde{K}'$\;

\BlankLine
\textbf{Transductive Kernel Classifier Steps}:

$T_{OVA} \leftarrow 2\cdot \boldsymbol{1}_c(T,:) - 1$\;

$i_{train} \leftarrow 1:m$\;

$i_{test} \leftarrow m+1:m+n$\;

\For{$s \in \{1,2\}$}
{
	$\ddot{K}_{train} \leftarrow \ddot{K}(i_{train},i_{train})$\;
	
	$\ddot{K}_{test} \leftarrow \ddot{K}(i_{test},i_{train})$\;
	
	$S_{OVA} \leftarrow \boldsymbol{0}_{n,c}$\;
	
	\For{$i \in \{1,2,...,c \}$}
	{
		$(\alpha, b) \leftarrow$ the dual weights of $\mathcal{C}$ trained on $\ddot{K}_{train}$ with the labels $T_{OVA}(:,i)$\;
	
		$S_{OVA}(:,i) \leftarrow \ddot{K}_{test} \cdot \alpha + b$\;
	}
	
	$P \leftarrow \boldsymbol{0}_{n,1}$;
	$S \leftarrow \boldsymbol{0}_{n,1}$\;
	
	\For{$i \in \{1,2,...,n \}$}
	{
		$P_i \leftarrow \mbox{argmax}(S_{OVA}(i,:))$\;

		$S_i \leftarrow \mbox{max}(S_{OVA}(i,:))$\;		
	}
	
	\If{$s = 1$}
	{
		$i_{sort} \leftarrow$ sort $S$ in descending order and return the sorted indexes\;
		      
		$i_{keep} \leftarrow i_{sort}(1:r)$\;
		
		$P_{keep} \leftarrow P(i_{keep})$\;
		
		$T \leftarrow T \cup P_{keep}$\;
		
		$T_{OVA} \leftarrow 2\cdot \boldsymbol{1}_c(T,:) - 1$\;
		
		$i_{train} \leftarrow i_{train} \cup i_{test}(i_{keep})$\;
	}
}

\BlankLine
\textbf{Output}: 

$P = \lbrace p_i \, \mid \, p_i \in \{1,2,...,c\}, i \in \{ 1,2, ..., n \} \rbrace$ -- the set of predicted labels for the test samples in $Y$.
}
\end{algorithm}

\vspace*{-0.1cm}
\section{Transductive Kernel Classifier}
\label{sec_TTL}
\vspace*{-0.1cm}

We next present a simple yet effective approach for adapting a one-versus-all kernel classifier trained on a source domain to a different target domain. Our transductive kernel classifier (TKC) approach is composed of two learning iterations. Our entire framework is formally described in Algorithm~\ref{alg_TTL}.

\noindent
{\bf Notations.}
We use the following notations in the algorithm. Sets, arrays and matrices are written in capital letters. All collection types are considered to be indexed starting from position $1$. The elements of a set $S$ are denoted by $s_i$, the elements of an array $A$ are alternatively denoted by $A(i)$ or $A_i$, and the elements of a matrix $M$ are denoted by $M(i,j)$ or $M_{ij}$ when convenient. The sequence $1,2,..., n$ is denoted by $1:n$. We use sequences to index arrays or matrices as well. For example, for an array $A$ and two integers $i$ and $j$, $A(i:j)$ denotes the sub-array $(A_i, A_{i+1}, ..., A_j)$. In a similar manner, $M(i:j,k:l)$ denotes a sub-matrix of the matrix $M$, while $M(i,:)$ returns the $i$-th row of M and $M(:,j)$ returns the $j$-th column of M. The zero matrix of $m \times n$ components is denoted by $\boldsymbol{0}_{m,n}$, and the square zero matrix is denoted by $\boldsymbol{0}_n$. The identity matrix is denoted by $\boldsymbol{1}_{n}$.

\noindent
{\bf Algorithm description.}
In steps $8$-$17$, we compute the domain-adapted string kernel matrix, as described in the previous section. In the first learning iteration (when $s = 1$), we train several classifiers to distinguish each individual class from the rest, according to the one-versus-all (OVA) scheme. In step $27$, the kernel classifier $\mathcal{C}$ is trained to distinguish a class from the others, assigning a dual weight to each training sample from the source domain. The returned column vector of dual weights is denoted by $\alpha$ and the bias value is denoted by $b$. The vector of weights $\alpha$ contains $m$ values, such that the weight $\alpha_i$ corresponds to the training sample $x_i$. When the test kernel matrix $\ddot{K}_{test}$ of $n \times m$ components is multiplied with the vector $\alpha$ in step $28$, the result is a column vector of $n$ positive or negative scores. Afterwards (step $34$), the test samples are sorted in order to maximize the probability of correctly predicted labels. For each test sample $y_i$, we consider the score $S_i$ (step $32$) produced by the classifier for the chosen class $P_i$ (step $31$), which is selected according to the OVA scheme. The sorting is based on the hypothesis that if the classifier associates a higher score to a test sample, it means that the classifier is more confident about the predicted label for the respective test sample. Before the second learning iteration, a number of $r$ test samples from the top of the sorted list are added to the training set (steps $35$-$39$) for another round of training. As the classifier is more confident about the predicted labels $P_{keep}$ of the added test samples, the chance of including noisy examples (with wrong labels) is minimized. On the other hand, the classifier has the opportunity to learn some useful domain-specific patterns of the test domain. %, as the majority of added test samples have correct labels. 
We believe that, at least in the cross-domain setting, the added test samples bring more useful information than noise. We would like to stress out that \emph{the ground-truth test labels are never used in our transductive algorithm}. Although the test samples are required beforehand, their labels are not necessary. Hence, our approach is suitable in situations where unlabeled data from the target domain can be collected cheaply, and such situations appear very often in practice, considering the great amount of data available on the Web.

\vspace*{-0.1cm} 
\section{Polarity Classification}
\label{sec_Polarity_Experiments}
\vspace*{-0.1cm} 

\noindent
{\bf Data set.}
For the cross-domain polarity classification experiments, we use the second version of Multi-Domain Sentiment Dataset \cite{Blitzer-ACL-2007}. The data set contains Amazon product reviews of four different domains: Books (B), DVDs (D), Electronics (E) and Kitchen appliances (K). Reviews contain star ratings (from 1 to 5) which are converted into binary labels as follows: reviews rated with more than 3 stars are labeled as positive, and those with less than 3 stars as negative. In each domain, there are 1000 positive and 1000 negative reviews.

\noindent
{\bf Baselines.}
We compare our approach with several methods \cite{Pan-WWW-2010,Bollegala-KDE-2013,Franco-KBS-2015,Sun-AAAI-2016,franco-EACL-2017,Huang-AAAI-2017} in two cross-domain settings. Using string kernels, Gim\'{e}nez-P\'{e}rez et al.~\cite{franco-EACL-2017} reported better performance than SST \cite{Bollegala-KDE-2013} and KE-Meta \cite {Franco-KBS-2015} in the multi-source domain setting. In addition, we compare our approach with SFA \cite{Pan-WWW-2010}, CORAL \cite{Sun-AAAI-2016} and TR-TrAdaBoost \cite{Huang-AAAI-2017} in the single-source setting. %Note that \cite{Pan-WWW-2010,Bollegala-KDE-2013,Franco-KBS-2015,Sun-AAAI-2016} report results in only one setting. %, while Gim\'{e}nez-P\'{e}rez et al.~\cite{franco-EACL-2017} report results in both settings.

\begin{table}[!t]
\setlength\tabcolsep{3.5pt}
\small{
\begin{center}
\begin{tabular}{lcccc}
\hline
% Method 															& Books			&	DVDs		&	Electronics		& Kitchen\\
Method 								& DEK$\rightarrow$B	&	BEK$\rightarrow$D	&	BDK$\rightarrow$E	& BDE$\rightarrow$K\\
\hline
SST~\cite{Bollegala-KDE-2013}			& $76.3$					& $78.3$					& $83.9$					& $85.2$ \\
KE-Meta~\cite{Franco-KBS-2015}     	& $77.9$					& $80.4$					& $78.9$					& $82.5$ \\
$K_{0/1}$~\cite{franco-EACL-2017}	& $82.0$ 					& $81.9$ 					& $83.6$ 					& $85.1$ \\
$K_{\cap}$~\cite{franco-EACL-2017}& $80.7$ 					& $80.7$ 					& $83.0$ 					& $85.2$ \\
\hline
\vspace{-0.9em}\\
$\ddot{K}_{0/1}$								& $82.9$					& $83.2$*					& $84.8$*					& $86.0$*\\
$\ddot{K}_{\cap}$								& $82.5$					& $82.9$*					& $84.5$*					& $86.1$*\\
$\ddot{K}_{0/1}$ + TKC 					& $\mathbf{84.1}$* 	& $\mathbf{84.0}$* 	& $\mathbf{85.4}$* 	& $86.9$*\\
$\ddot{K}_{\cap}$ + TKC	& 				$83.8$*					& $83.5$*					& $85.0$*					& $\mathbf{87.1}$*\\
\hline
\end{tabular}
\end{center}
}
%\vspace*{-0.1cm}
\caption{Multi-source cross-domain polarity classification accuracy rates (in $\%$) of our transductive approaches versus a state-of-the-art baseline based on string kernels \cite{franco-EACL-2017}, as well as SST \cite{Bollegala-KDE-2013} and KE-Meta \cite{Franco-KBS-2015}. The best accuracy rates are highlighted in bold. The marker * indicates that the performance is significantly better than the best baseline string kernel according to a paired McNemar's test performed at a significance level of $0.01$.}
\label{tab_Polarity_Multi}
\vspace*{-0.3cm}
\end{table}

\noindent
{\bf Evaluation procedure and parameters.}
We follow the same evaluation methodology of Gim\'{e}nez-P\'{e}rez et al.~\cite{franco-EACL-2017}, to ensure a fair comparison. Furthermore, we use the same kernels, namely the presence bits string kernel ($K_{0/1}$) and the intersection string kernel ($K_{\cap}$), and the same range of character n-grams (5-8). To compute the string kernels, we used the open-source code provided by Ionescu et al.~\cite{ionescu-popescu-cahill-EMNLP-2014, radu-marius-book-chap6-2016}.
For the transductive kernel classifier, we select $r=1000$ unlabeled test samples to be included in the training set for the second round of training. We choose Kernel Ridge Regression \cite{taylor-Cristianini-cup-2004} as classifier and set its regularization parameter to $10^{-5}$ in all our experiments. Although Gim\'{e}nez-P\'{e}rez et al.~\cite{franco-EACL-2017} used a different classifier, namely Kernel Discriminant Analysis, we observed that Kernel Ridge Regression produces similar results ($\pm 0.1\%$) when we employ the same string kernels.
As Gim\'{e}nez-P\'{e}rez et al.~\cite{franco-EACL-2017}, we evaluate our approach in two cross-domain settings. In the multi-source setting, we train the models on all domains, except the one used for testing. In the single-source setting, we train the models on one of the four domains and we independently test the models on the remaining three domains.

\begin{table*}[!th]
\setlength\tabcolsep{4.5pt}
\small{
\begin{center}
\begin{tabular}{lcccccccccccc}
\hline
Method 												& D$\rightarrow$B			& E$\rightarrow$B		& K$\rightarrow$B		
															& B$\rightarrow$D 			& E$\rightarrow$D 		& K$\rightarrow$D\\
\hline
SFA~\cite{Pan-WWW-2010}					& $79.8$ 						& $78.3$ 					& $75.2$ 
															& $81.4$ 						& $77.2$ 					& $\mathbf{78.5}$\\
															
CORAL~\cite{Sun-AAAI-2016}				& $78.3$ 						& - 							& - 
															& - 								& - 							& $73.9$\\
															
TR-TrAdaBoost~\cite{Huang-AAAI-2017}& $74.7$ 					& $69.1$ 					& $70.6$ 
															& $79.6$ 						& $71.8$ 					& $74.4$\\
															
$K_{0/1}$~\cite{franco-EACL-2017}		& $82.0$ 						& $72.4$ 					& $72.7$ 
															& $81.4$ 						& $74.9$ 					& $73.6$\\
																
$K_{\cap}$~\cite{franco-EACL-2017}	& $82.1$ 						& $72.4$ 					& $72.8$ 
															& $81.3$ 						& $75.1$					& $72.9$\\
\hline
\vspace{-0.9em}\\
$\ddot{K}_{0/1}$									& $83.3$* 					& $74.5$* 					& $74.3$*
															& $83.0$* 					& $76.9$* 					& $74.9$*\\
																					
$\ddot{K}_{\cap}$									& $83.2$* 					& $74.2$* 					& $74.0$*
															& $82.8$* 					& $76.4$* 					& $75.1$*\\
																					
$\ddot{K}_{0/1}$ + TKC 						& $\mathbf{84.9}$* 		& $\mathbf{78.5}$* 		& $\mathbf{76.6}$*
															& $84.0$* 					& $\mathbf{79.6}$* 		& $76.4$*\\
																					
$\ddot{K}_{\cap}$ + TKC						& $84.5$* 					& $\mathbf{78.5}$* 		& $75.8$*				
															& $\mathbf{84.2}$* 		& $79.1$* 					& $76.5$*\\
\hline
\hline
Method													& B$\rightarrow$E 			& D$\rightarrow$E 		& K$\rightarrow$E
															& B$\rightarrow$K 			& D$\rightarrow$K 		& E$\rightarrow$K\\
\hline
SFA~\cite{Pan-WWW-2010}					& $73.5$ 						& $76.7$ 					& $85.1$ 
															& $79.1$ 						& $80.8$ 					& $86.8$\\
															
CORAL~\cite{Sun-AAAI-2016}				& $76.3$ 						& - 							& - 
															& - 								& - 							& $83.6$\\
															
TR-TrAdaBoost~\cite{Huang-AAAI-2017}& $74.9$ 						& $75.9$ 					& $83.1$ 
															& $77.8$ 						& $75.7$ 					& $83.7$\\
															
$K_{0/1}$~\cite{franco-EACL-2017}		& $71.3$ 						& $74.4$ 					& $83.9$ 
															& $74.6$ 						& $75.4$ 					& $84.9$\\
																
$K_{\cap}$~\cite{franco-EACL-2017}	& $71.8$ 						& $74.5$ 					& $84.4$
															& $74.9$ 						& $75.1$ 					& $84.9$\\
\hline
\vspace{-0.9em}\\															
$\ddot{K}_{0/1}$									& $74.0$* 					& $76.0$* 					& $85.4$*
															& $77.6$* 					& $77.3$* 					& $86.0$*\\
																					
$\ddot{K}_{\cap}$									& $74.2$* 					& $75.9$* 					& $85.2$*
															& $77.6$* 					& $77.3$* 					& $85.9$*\\
																					
$\ddot{K}_{0/1}$ + TKC 						& $76.6$* 					& $\mathbf{77.1}$* 		& $\mathbf{86.4}$*
															& $\mathbf{79.6}$* 		& $\mathbf{80.9}$* 		& $\mathbf{87.0}$*\\
																					
$\ddot{K}_{\cap}$ + TKC						& $\mathbf{76.7}$* 		& $76.8$* 					& $\mathbf{86.4}$*
															& $79.4$* 					& $80.5$* 					& $\mathbf{87.0}$*\\
\hline
\end{tabular}
\end{center}
}
%\vspace*{-0.1cm}
\caption{Single-source cross-domain polarity classification accuracy rates (in $\%$) of our transductive approaches versus a state-of-the-art baseline based on string kernels \cite{franco-EACL-2017}, as well as SFA \cite{Pan-WWW-2010}, CORAL \cite{Sun-AAAI-2016} and TR-TrAdaBoost \cite{Huang-AAAI-2017}. The best accuracy rates are highlighted in bold. The marker * indicates that the performance is significantly better than the best baseline string kernel according to a paired McNemar's test performed at a significance level of $0.01$.}
\label{tab_Polarity_Single}
\vspace*{-0.3cm}
\end{table*}

\noindent
{\bf Results in multi-source setting.}
The results for the multi-source cross-domain polarity classification setting are presented in Table~\ref{tab_Polarity_Multi}. Both the transductive presence bits string kernel ($\ddot{K}_{0/1}$) and the transductive intersection kernel ($\ddot{K}_{\cap}$) obtain better results than their original counterparts. %For instance, on the Electronics target domain the accuracy of the presence bits kernel grows from $83.6\%$ to $84.8\%$ for the domain-adapted version, while the accuracy of the intersection kernel grows from $83.0\%$ to $84.5\%$. 
Moreover, according to the McNemar's test~\cite{Dietterich-NC-1998}, the results on the DVDs, the Electronics and the Kitchen target domains are significantly better than the best baseline string kernel, with a confidence level of $0.01$. 
When we employ the transductive kernel classifier (TKC), we obtain even better results. On all domains, the accuracy rates yielded by the transductive classifier are more than $1.5\%$ better than the best baseline. For example, on the Books domain the accuracy of the transductive classifier based on the presence bits kernel ($84.1\%$) is $2.1\%$ above the best baseline ($82.0\%$) represented by the intersection string kernel. Remarkably, the improvements brought by our transductive string kernel approach are statistically significant in all domains.

\noindent
{\bf Results in single-source setting.} The results for the single-source cross-domain polarity classification setting are presented in Table~\ref{tab_Polarity_Single}. We considered all possible combinations of source and target domains in this experiment, and we improve the results in each and every case. Without exception, the accuracy rates reached by the transductive string kernels are significantly better than the best baseline string kernel \cite{franco-EACL-2017}, according to the McNemar's test performed at a confidence level of $0.01$. The highest improvements (above $2.7\%$) are obtained when the source domain contains Books reviews and the target domain contains Kitchen reviews. %In this case (Books$\rightarrow$Kitcken), for instance, the accuracy of the presence bits kernel grows by $3.0\%$, from $74.6\%$ to $77.6\%$, for the domain-adapted version.
As in the multi-source setting, we obtain much better results when the transductive classifier is employed for the learning task. In all cases, the accuracy rates of the transductive classifier are more than $2\%$ better than the best baseline string kernel. Remarkably, in four cases (E$\rightarrow$B, E$\rightarrow$D, B$\rightarrow$K and D$\rightarrow$K) our improvements are greater than $4\%$. %For example, when the source domain contains Electronics reviews, the target domain contains Books reviews, and the TTL classifier is used for training, both the domain-adapted presence bits string kernel and the domain-adapted intersection kernel exhibit an improvement of $6.1\%$ (from $72.4\%$ to $78.5\%$) over their original counterparts.
The improvements brought by our transductive classifier based on string kernels are statistically significant in each and every case. In comparison with SFA \cite{Pan-WWW-2010}, we obtain better results in all but one case (K$\rightarrow$D). Remarkably, we surpass the other state-of-the-art approaches \cite{Sun-AAAI-2016,Huang-AAAI-2017} in all cases.

\vspace*{-0.1cm} 
\section{Conclusion}
\label{sec_Conclusion}
\vspace*{-0.1cm} 

In this paper, we presented two domain adaptation approaches that can be used together to improve the results of string kernels in cross-domain settings. We provided empirical evidence indicating that our framework can be successfully applied in cross-domain text classification, particularly in cross-domain English polarity classification. Indeed, the polarity classification experiments demonstrate that our framework achieves better accuracy rates than other state-of-the-art methods \cite{Pan-WWW-2010,Bollegala-KDE-2013,Franco-KBS-2015,Sun-AAAI-2016,franco-EACL-2017,Huang-AAAI-2017}. By using the same parameters across all the experiments, we showed that our transductive transfer learning framework can bring significant improvements without having to fine-tune the parameters for each individual setting.
Although the framework described in this paper can be generally applied to any kernel method, we focused our work only on string kernel approaches used in text classification. 
In future work, we aim to combine the proposed transductive transfer learning framework with different kinds of kernels and classifiers, and employ it for other cross-domain tasks.

%\vspace{-0,1cm}
\bibliography{references}{} 
\bibliographystyle{splncs03}

\end{document}